\documentclass[3p]{elsarticle}

\usepackage{lineno,hyperref}

\journal{Journal of \LaTeX\ Templates}

\biboptions{authoryear}

\usepackage{booktabs}       
\usepackage{subcaption}
\usepackage{amsmath}
\usepackage{graphicx}
\usepackage{wrapfig}
\usepackage{amsfonts}
\usepackage{multirow}

\begin{document}

\begin{frontmatter}

\title{A Compact Network Learning Model for Distribution Regression\textsl{}}


\author[soc]{Connie Khor Li Kou \corref{cor1}}
\ead{koukl@comp.nus.edu.sg}
\author[astar,soc]{Hwee Kuan Lee}
\ead{leehk@bii.a-star.edu.sg}
\author[soc]{Teck Khim Ng}
\ead{ngtk@comp.nus.edu.sg}

\cortext[cor1]{Corresponding author}

\address[soc]{School of Computing, National University of Singapore, 13 Computing Drive, Singapore 117417}
\address[astar]{Bioinformatics Institute, Agency for Science, Technology and Research (A*STAR), 30 Biopolis Street, Singapore 138671}

\begin{abstract}
Despite the superior performance of deep learning in many applications, challenges remain in the area of regression on function spaces. In particular, neural networks are unable to encode function inputs compactly as each node encodes just a real value. We propose a novel idea to address this shortcoming: to encode an entire function in a single network node. To that end, we design a compact network representation that encodes and propagates functions in single nodes for the distribution regression task. Our proposed Distribution Regression Network (DRN) achieves higher prediction accuracies while being much more compact and uses fewer parameters than traditional neural networks.
\end{abstract}

\begin{keyword}
Supervised Learning \sep Distribution Regression
\end{keyword}

\end{frontmatter}


\section{Introduction}
Deep neural networks have achieved state-of-the-art results in many machine learning tasks. By constructing deep and complex architectures from simple units, deep networks have the ability to perform difficult prediction tasks well. However, this manner of network construction results in a large number of parameters, which leads to problems such as difficulties in optimization, overfitting and large memory usage. Indeed, these problems have motivated many research efforts in deep learning. For instance, the use of shared weights and pooling layers in convolutional neural networks reduces the optimization search space and alleviates overfitting \citep{lecun1989backpropagation,lin2013network}. Regularization techniques like dropout and DropConnect are used to prevent overfitting in large networks \citep{srivastava2014dropout,wan2013regularization}. Furthermore, numerous network compression techniques have been proposed to aid deployment of neural nets in low-memory devices \citep{han2015deep, novikov2015tensorizing}.

In spite of these efforts on reducing number of parameters, challenges remain in the area of regression on function spaces: $f: \mathcal{G}\to \mathcal{H}$, where $ \mathcal{G}$ and  $\mathcal{H}$ are function spaces ($g \in \mathcal{G}$, $h \in \mathcal{H}$, where $g:  \mathbb{R} \to \mathbb{R}$ and $h:  \mathbb{R} \to \mathbb{R}$) and $f$ is the regression from $\mathcal{G}$ to $\mathcal{H}$. The challenge lies in encoding the functions $g$ and $h$ in a neural network. One method is to sample the function with a discretized grid of points, while another is to use basis functions approximation \citep{oliva2015fast}. In any case, the function has to be broken into smaller parts before feeding as inputs to the network. In neural networks, each node only encodes a real value. Hence to represent functions, we require many input nodes and weights, resulting in large parameter sizes. Instead of decomposing a function into different nodes, we propose a novel idea: to use a single node to encode the entire function.

\begin{figure}[]
	\centering
	\includegraphics[width=0.9\linewidth]{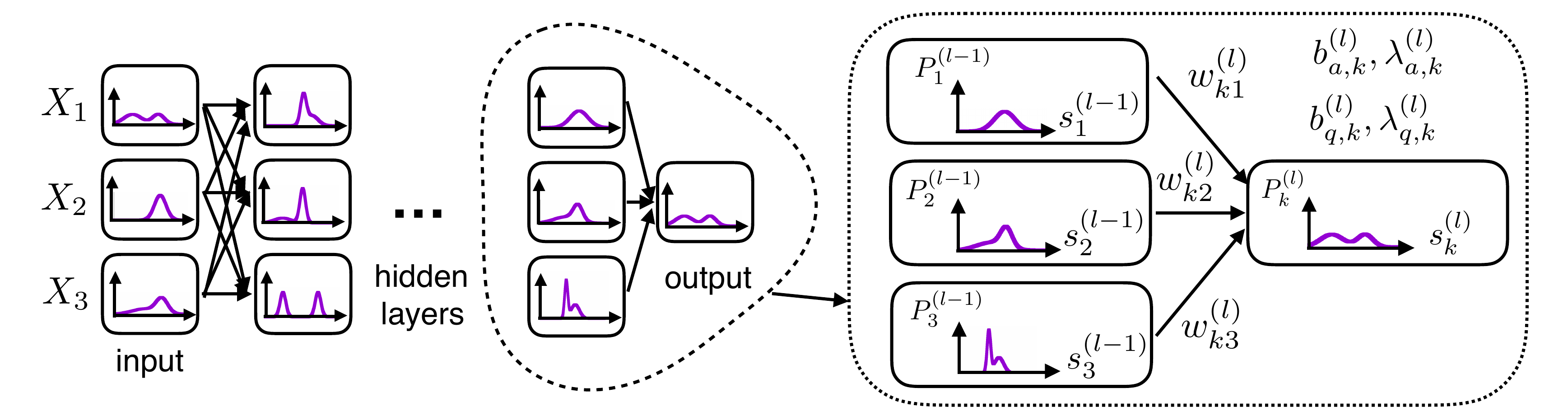}
	\caption{(Left) An example DRN with multiple input distributions mapping to an output distribution. (Right) A connection unit in the network, with 3 input nodes in layer $l-1$ connecting to a node in layer $l$. Each node encodes a probability distribution. The tunable parameters are the connecting weights and the bias parameters at the output node.}
	\label{fig:forwardprop_and_nw}
\end{figure}

In this paper, we propose a network design inspired by statistical physics with the use of harmonic energy functions and Boltzmann distributions. Indeed, as we show later, our network representation is very compact and it models the Ornstein-Uhlenbeck stochastic process with just one node in the hidden layer. We use this network specifically for the distribution regression task where we regress from input distributions to output distributions. This idea is illustrated in Fig. \ref{fig:forwardprop_and_nw}. Our network, named Distribution Regression Network (DRN), has a more compact representation than traditional neural nets such as multilayer perceptrons (MLP) and convolutional neural networks (CNN). On real-world datasets, DRN outperforms neural nets while using much fewer parameters. We also demonstrate the advantages of our network by comparing with other state-of-the-art distribution regression methods \citep{oliva2015fast}.

\section{Related work}
Regression on distributions has many relevant applications. In the study of human populations, applications include predicting voting outcomes of demographic groups \citep{flaxman2016understanding} and predicting economic growth from income distribution \citep{perotti1996growth}. In particular, distribution-to-distribution regression is useful in predicting outcomes of phenomena driven by stochastic processes. For example, the Ornstein-Uhlenbeck diffusion process is used to model commodity prices \citep{schwartz2000short} and phenotypic traits in quantitative biology \citep{bartoszek2016ornstein}.

Variants of the distribution regression task have been explored in literature \citep{poczos2013distribution,oliva2014fast}. For the distribution-to-distribution regression task, \citet{lampert2015predicting} proposed the Extrapolating the Distribution Dynamics method which predicts the future state of a time-varying  distribution given a sequence of samples from previous time steps. However, it is unclear how it can be used for regressing distributions of different objects. For general distribution regression, \citet{oliva2013distribution} proposed an instance-based learning method where a linear smoother estimator (LSE) is applied across the data. However, the computation time of LSE scales badly with the size of the dataset.  \citet{oliva2015fast} developed the Triple-Basis Estimator (3BE) for function-to-function regression, where the prediction time is independent of the number of data by using basis representations of functions and Random Kitchen Sink (RKS) basis functions. The authors have extended 3BE for distribution regression, showing improved accuracy and speed compared to LSE. Hence, 3BE is considered state-of-the-art for distribution-to-distribution regression and we compare with 3BE and other neural network architectures in our experiments.

The notion of mapping from a distribution to another is seen in generative models such as generative adversarial networks and variational autoencoders. In such models, the network learns to map a single latent distribution to a single data distribution \citep{goodfellow2014generative,kingma2013auto,dinh2016density}. However, we emphasize this is different from our supervised distribution regression task as we learn a mapping from multiple pairs of input-output distributions.

\section{Distribution Regression Network} \label{sect:DRN}
Given a training dataset with $M$ data points $\mathcal{D} = \{(X_1^1,\cdots,X_1^K,Y_1), \cdots, (X_M^1,\cdots,X_M^K,Y_M)\}$ where $X_i^k$ and $Y_i$ are univariate continuous distributions with compact support, the regression task is to learn the function $f$ which maps the input distributions to output distribution: $Y_i=f(X_i^1,\cdots,X_i^K)$. It is trivial to generalize our method to multiple output distributions but for simplicity of explanation we shall restrict to single output regressions in the discussions.
\subsection{Forward propagation} \label{sect:forwardprop}
Fig. \ref{fig:forwardprop_and_nw} illustrates how the regression is realized. DRN adapts the conventional neural network by encoding an entire probability distribution in a single node. Each layer of nodes is connected to the next layer by real-valued weights. The input consists of one or more probability distributions which are propagated layerwise through the hidden layers. Even though each node encodes a probability distribution, DRN is different from a bayes net. Unlike bayes net where the conditional probability among variables are learnt by maximizing the likelihood over observed data, DRN regresses distributions using a feedforward network.

At each node in the hidden layer, the probability distribution is computed using the distributions of the incoming nodes in the previous layer and the weights and bias parameters. $P_k^{(l)}$ represents the probability density function (pdf) of the $k^{\text{th}}$ node in the $l^{\text{th}}$ layer and $P_k^{(l)}(s_k^{(l)})$ is the density of the pdf when the node's random variable is $s_k^{(l)}$. Before obtaining the probability distribution $P_k^{(l)}$, we first compute its unnormalized form $\tilde{P}_k^{(l)}$.  $\tilde{P}_k^{(l)}$ is computed by marginalizing over the product of the unnormalized conditional probability $\tilde{Q}(s_k^{(l)}|s_1^{(l-1)},\cdots,s_n^{(l-1)})$ and the incoming probabilities.
\begin{align}
\label{eq:margin}
\tilde{P}_k^{(l)}\left(s_k^{(l)}\right) = \int_{{s_1}^{(l-1)}}  \cdots \int_{{s_n}^{(l-1)}}
&\tilde{Q}\left(s_k^{(l)}|s_1^{(l-1)},\cdots,s_n^{(l-1)}\right) \\ \nonumber
& P_1^{(l-1)}\left(s_1^{(l-1)}\right)\cdots P_n^{(l-1)}\left(s_n^{(l-1)}\right)  \,ds_1^{(l-1)} \cdots ds_n^{(l-1)}
\end{align}
\begin{align}
\label{eq:condprob}
\tilde{Q}\left(s_k^{(l)}|s_1^{(l-1)},\cdots,s_n^{(l-1)}\right) = \exp\left[ -E\left(s_k^{(l)}|s_1^{(l-1)},\cdots,s_n^{(l-1)}\right)\right]
\end{align}
$s_1^{(l-1)},\cdots,s_n^{(l-1)}$ represent the variables of the lower layer nodes and $E$ is the energy for a given set of node variables, which we define later in Eq. (\ref{eq:energy}). The unnormalized conditional probability has the same form as the Boltzmann distribution, except that the partition function is omitted. This omission reduces the computational complexity of our model through factorization, shown later in Eq. (\ref{eq:factorizeP}). 

Our energy function formulation is motivated by work on spin models in statistical physics where spin alignment to coupling fields and critical phenomena are studied \citep{katsura1962statistical, lee2002monte, lee2003monte, wu1982potts}. Energy functions are also used in other network models. In the energy-based models of \citet{teh2003energy} and \citet{lecun2007energy}, the parameters are learnt such that the observed configurations of the variables have lower energies than unobserved ones. Our DRN is different as the energy function is part of the forward propagation process and is not directly optimized. For a given set of node variables, the energy function in DRN is
\begin{align}
\label{eq:energy}
E\left(s_k^{(l)}|s_1^{(l-1)},\cdots,s_n^{(l-1)}\right)  = \sum_i^n w_{ki}^{(l)}
\left( \frac{s_k^{(l)}-s_i^{(l-1)}}{\Delta}\right)^2 
&+ b_{q,k}^{(l)}\left( \frac{s_k^{(l)}-\lambda_{q,k}^{(l)}}{\Delta} \right)^2  + b_{a,k}^{(l)} \left|\frac{s_k^{(l)}-\lambda_{a,k}^{(l)}}{\Delta} \right|. 
\end{align}
$w_{ki}^{(l)}$ is the weight connecting the $i^\text{th}$ node in layer $l-1$ to the $k^{th}$ node in layer $l$.  $b_{q,k}^{(l)}$ and $b_{a,k}^{(l)}$ are the values of the quadratic and absolute bias terms which act at the positions $\lambda_{q,k}^{(l)}$ and $\lambda_{a,k}^{(l)}$ respectively. The purpose of the individual terms in Eq. \ref{eq:energy} is discussed in detail in Section \ref{sect:behave} and Fig. \ref{fig:propbehave}. The support length of the distribution is given by $\Delta$. All terms in Eq. (\ref{eq:energy}) are normalized so that the energy function is invariant to the support length. Eq. (\ref{eq:margin}) can be factorized such that instead of having multidimensional integrals, there are $n$ univariate integrals:
\begin{align}
\label{eq:factorizeP}
&\tilde{P}_k^{(l)} \left(s_k^{(l)}\right) =  e^{-B\left(s_k^{(l)}\right) }  \prod_i^n  \int\limits_{s_i^{(l-1)}} \!\!\! P_i^{(l-1)}\left(s_i^{(l-1)}\right) e^{ - w_{ki}^{(l)} \left( \frac{s_k^{(l)}-s_i^{(l-1)}}{\Delta}\right)^2 }  \,ds_i ^{(l-1)} ,
\end{align}
where $B(s_k^{(l)})$ captures the bias terms in Eq. (\ref{eq:energy}).
\begin{align}
B\left(s_k^{(l)}\right) = b_{q,k}^{(l)}\left( \frac{s_k^{(l)}-\lambda_{q,k}^{(l)}}{\Delta} \right)^2  + b_{a,k}^{(l)} \left|\frac{s_k^{(l)}-\lambda_{a,k}^{(l)}}{\Delta} \right| 
\label{eq:B}
\end{align}
Finally, the distribution from Eq. (\ref{eq:margin}) is normalized. 
\begin{align}
P_k^{(l)}\left(s_k^{(l)}\right) =\frac{ \tilde{P}_k^{(l)}\left(s_k^{(l)}\right) }{\int_{s_k^{(l)'}} \tilde{P}_k^{(l)}\left(s_k^{(l)'}\right) \, ds_k^{(l)'}}
\label{eq:norm}
\end{align}
The propagation of probability distributions within a connection unit forms the basis for forward propagation. Forward propagation is performed layer-wise from the inputs to obtain the prediction at the final layer. 
\subsubsection{Propagation behavior}\label{sect:behave}
\begin{figure}
	\centering
	\begin{subfigure}[b]{0.64\columnwidth}
		\includegraphics[width=\columnwidth]{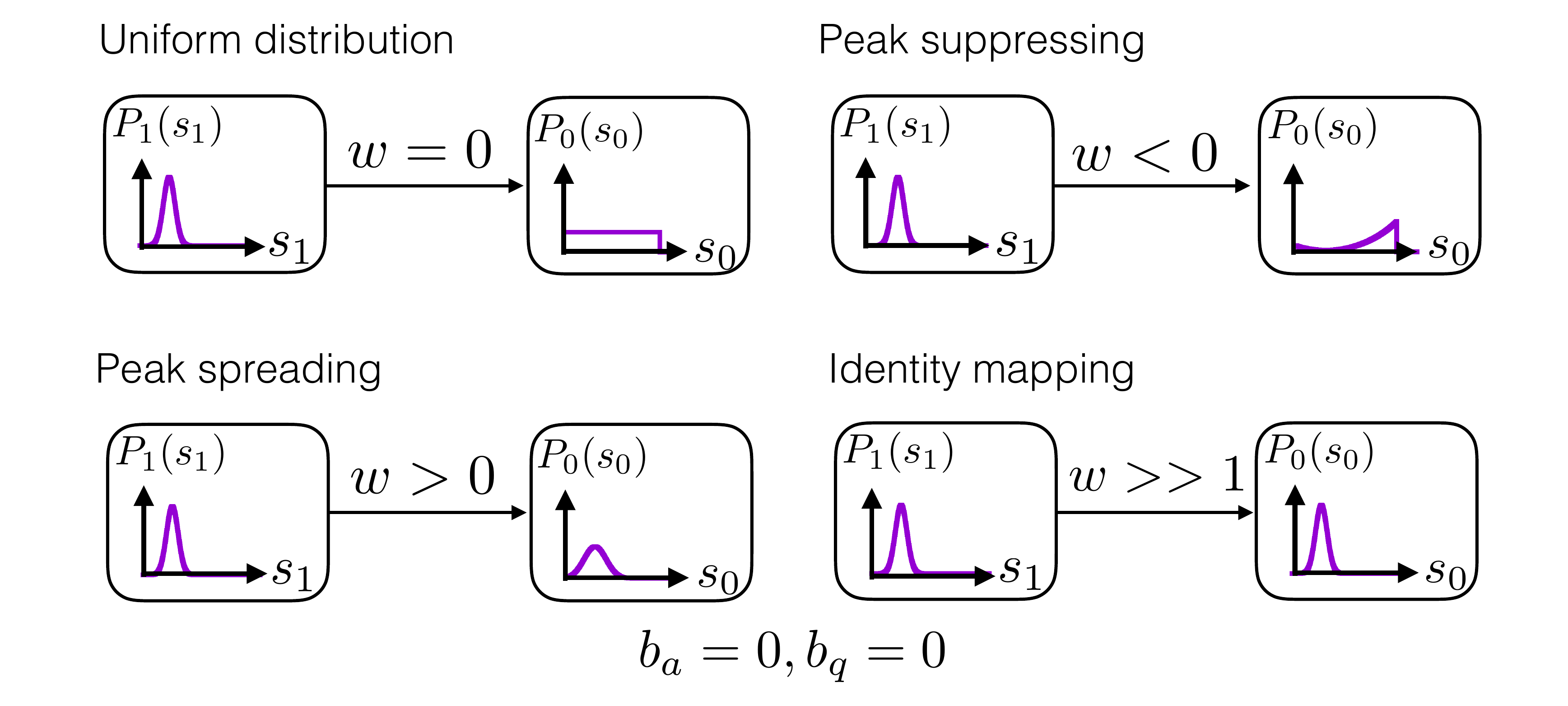}
		\caption{ }
		\label{fig:propbehave_weights}
	\end{subfigure}
	\quad
	\begin{subfigure}[b]{0.32\columnwidth}
		\includegraphics[width=\columnwidth]{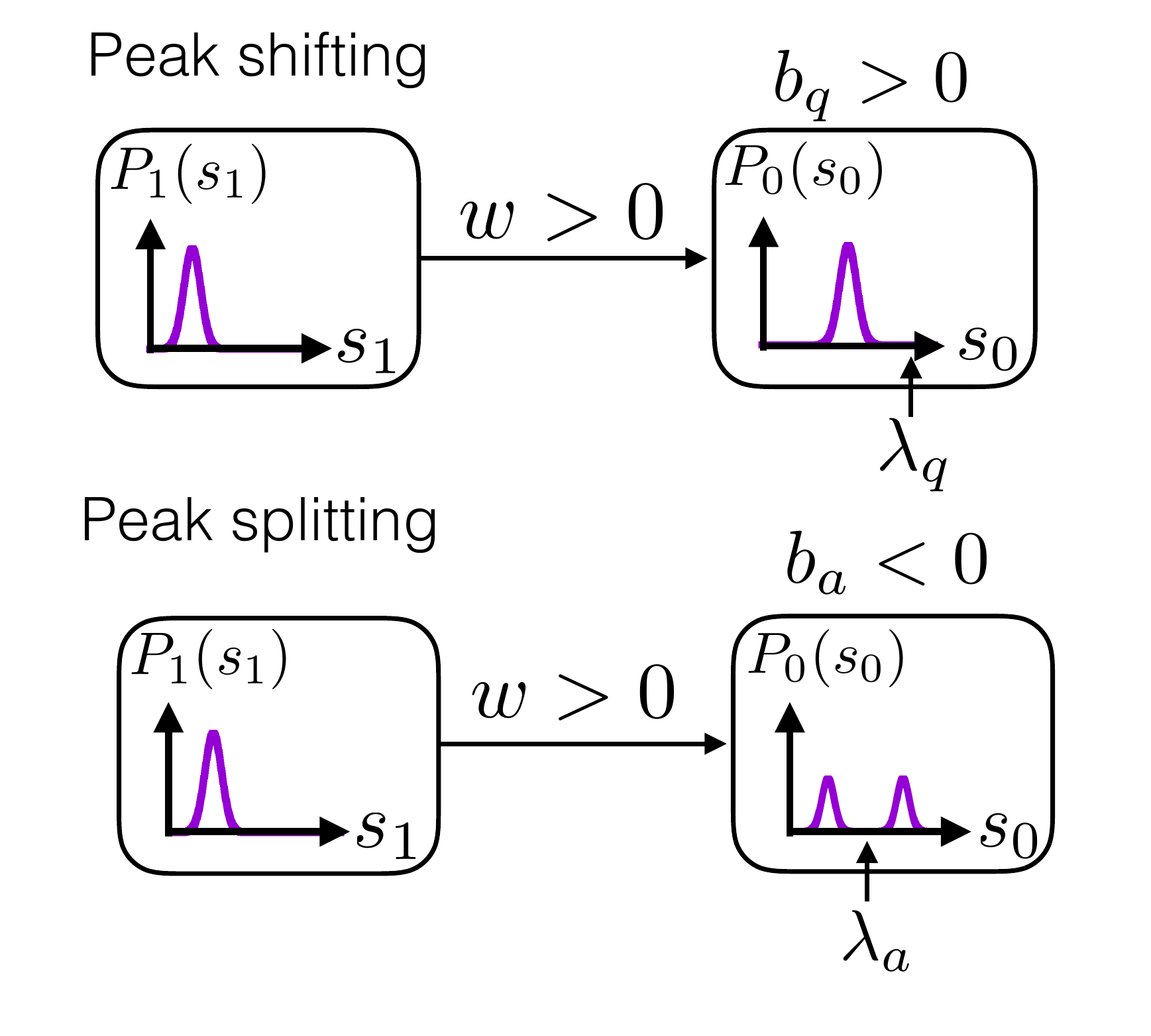}
		\caption{}
		\label{fig:propbehave_bias}
	\end{subfigure}
	\caption{Propagation behavior: (a)When weight is zero, the output distribution is flat. Positive weights cause the output to have the same peak position while negative weights cause the output to be lowest at the input peak position. When the weight is a sufficiently large positive number, the propagation tends towards the identity. (b) A positive quadratic bias shifts the input peak towards the bias position. A negative absolute bias splits the peak into two, centered about the bias position.}
	\label{fig:propbehave}
\end{figure}
We design the DRN model to exhibit important properties required to perform distribution regression tasks. Fig. \ref{fig:propbehave_weights} illustrates the propagation behavior for a connection unit with one input node where the bias values are set as zero. When the weight is zero, the output distribution is flat. Since there is no influence from the input distribution or the bias terms, all output variables are equally likely. This corresponds to the principle of maximum entropy in information theory.

With a positive weight, the output distribution has the same peak position as the input distribution whereas a negative weight causes the output distribution to be lowest at the input peak position. Such properties are important in capturing the correlation between the input and output distributions. In addition, the weight magnitude represents the strength of the effect. When the weight is a sufficiently large positive number, the propagation tends towards the identity mapping as the input distribution has full influence on the output. The implication is that like in neural networks, a deeper network should have at least the same complexity as a shallow one, as the added layers can produce the identity function. Conversely, a small positive weight causes the output peak to be at the same position as the input peak, but with more spread, hence it is able to model diffusive processes. The propagation behavior is a consequence of DRN's propagation formulation. The Boltzmann distribution in Eq. (\ref{eq:condprob}) coupled with the quadratic form of the weight term in Eq. (\ref{eq:energy}) causes the conditional probability to follow a Gaussian function, where the magnitude of a positive weight controls the spread.

The remaining quadratic and absolute bias terms play a similar role as the bias in traditional neural networks. The absolute and quadratic bias terms, as controlled by $b_{a,k}^{(l)}$ and $b_{q,k}^{(l)}$, act at the positions $\lambda_{a,k}^{(l)}$ and $\lambda_{q,k}^{(l)}$ respectively. The absolute and quadratic bias terms have different mathematical forms which give rise to the respective propagation behaviors, as shown in Fig. \ref{fig:propbehave_bias}. The quadratic bias is used to shift the input distribution towards the bias position. In the energy function, the quadratic term from the bias is added to the quadratic term from the weight, the resulting function, which is also quadratic, is shifted towards the bias location. 

While the quadratic bias term is able to shift the input peak, it is unable to give rise to bifurcations since the summation of quadratic functions is a quadratic function with only one stationary point. For this, we introduce the absolute bias. A summation of an inverted absolute function and a quadratic function can exhibit two stationary points in the energy function. Hence, a negative absolute bias can split an input peak into two, centered about the bias position (see Fig. \ref{fig:propbehave_bias}). The ability to model bifurcations, along with its simple mathematical form, makes the absolute bias a useful term in our formulation. Since shifting and bifurcation of distributions are common in many real-life phenomena, it is important to have both absolute and quadratic bias terms in our model.
\subsection{Network cost function}
The cost function of the network given a set of network parameters is measured by the Jensen-Shannon (JS) divergence \citep{lin1991divergence} between the label ($Y_i$) and predicted ($\hat{Y}_i$) distributions, denoted here as $D_{JS}(Y_i||\hat{Y}_i) $. The JS divergence is a suitable cost function as it is symmetric and bounded. The network cost function $C_{net}$ is the average $D_{JS}$ over all $M$ training data: $C_{net} = \frac{1}{M} \sum_i^M D_{JS}(Y_i||\hat{Y_i})$.
\subsection{Discretization of probability distributions}
In our experiments, the integrals in Eq. (\ref{eq:factorizeP}) are performed numerically. This is done through discretization from continuous probability density functions (pdf) to discrete probability mass functions (pmf). Given a continuous pdf with finite support, the range of the continuous variable is partitioned into $q$ equal widths and the probability distribution is binned into $q$ states. The estimation error arising from the discretization step will decrease with larger $q$.
\subsection{Optimization by backpropagation}\label{sect:opt}
The network cost is differentiable over the parameters. Similar to backpropagation, we derive at each node a $q$-by-$q$ matrix which denotes the derivative of the final layer node distribution with respect to the current node distribution:
\begin{align}
\frac{\partial P^{(L)}_1 (s^{(L)}_1)}{\partial P^{(l)}_k (s^{(l)}_k)} = \sum_i^n \sum_{s^{(l+1)}_i} \frac{\partial P^{(L)}_1 (s^{(L)}_1)}{\partial P^{(l+1)}_i (s^{(l+1)}_i)} \frac{\partial P^{(l+1)}_i (s^{(l+1)}_i)}{\partial P^{(l)}_k (s^{(l)}_k)},
\label{eq:backprop}
\end{align}
where $P^{(L)}_1 (s^{(L)}_1)$ is the final layer output distribution. From the derivative $\partial P^{(L)}_1 (s^{(L)}_1) / \partial P^{(l)}_k (s^{(l)}_k) $, the cost gradients for all network parameters can be obtained. Detailed derivations of the cost gradients are in  \ref{sect:appen_costgrad}. The network weights and bias magnitudes are randomly initialized with a uniform distribution, though other initialization methods are also feasible. The bias positions are uniformly sampled from the range corresponding to the support of the distributions.

\section{Experiments}\label{sect:expt}
In the following experiments we evaluate DRN on synthetic and real-world datasets. For the benchmark MLP architectures, we explored two ways of representing a distribution - by discretizing the bins and by decomposition into sinusoidal basis coefficients. We compared with 1) MLP with discretized bins as inputs (MLP-bins), 2) MLP with sinusoidal basis coefficients as inputs (MLP-basis), 3) one-dimensional convolutional neural net (1D-CNN) with discretized bins as inputs and 4) 3BE \citep{oliva2015fast}.  3BE is chosen as it is state-of-the-art for distribution regression, and we benchmark with MLP and 1D-CNN to study the advantages of encoding a distribution in each node. In particular, we chose 1D-CNN as it captures the continuity and locality in distributions through shared weights.

For MLP-bins and 1D-CNN, each distribution is represented by $q$ nodes. MLP-bins consists of fully connected layers and a softmax final layer. 1D-CNN consists of convolutional layers, fully-connected layers and a softmax final layer. MLP-bins and 1D-CNN are optimized with mean squared error; we also tried JS divergence but it gave worse performance. For MLP-basis and 3BE, each distribution is represented by its sinusoidal basis coefficients, where the number of basis was chosen by cross-validation. MLP-basis has fully-connected layers with a final linear layer for the basis coefficients of the output distribution, and the cost function is the mean squared error. In all our experiments, we conduct cross-validation for hyperparameter tuning and to avoid overfitting. In \ref{sect:appen_exptdetails}, we include the details of the experimental setup (eg. number of data, discretization size) and model architectures.

\begin{table}[]
	\centering
	\small
	\begin{tabular}{@{}ccccccccc@{}}  \hline	
		& \multicolumn{2}{c}{OU} & \multicolumn{2}{c}{Fokker-Planck} & \multicolumn{2}{c}{Cell} & \multicolumn{2}{c}{Stock} \\
		& $L2 (\times10^{-2})$    & $N_{param}$  & $NLL$         & $N_{param}$       & $NLL$    & $N_{param}$   & $NLL$     & $N_{param}$   \\  \hline	
		DRN       & $\mathbf{3.8(0.1)}$   & 10   & $\mathbf{-1291.8(0.2)}$     & 280               & $\mathbf{-148.5(0.4)}$   & 5             & $\mathbf{-474.4(0.1)}$    & 7             \\
		MLP-bins  & $5.2(0.2)$   & 5800         & $-1290.7(0.5)$       & 4960              & $-147.8(0.1)$   & 4120          & $-471.5(0.1)$    & 4110          \\
		MLP-basis & $6.0(0.3)$   & 900          &    $-1291.6(1.3)$    &    302    & $-145.7(0.4)$  & 162           & $-459.8(3.3)$    & 350           \\
		1D-CNN    & $4.4(0.2)$   & 1140         & $-1291.5(0.2)$       & 4620              & $-146.2(0.2)$   & 381           & $-471.5(0.1)$    & 1415          \\
		3BE       & $4.5(0.3)$  & 8800         & $-1288.3(3.2)$       & 95000             & $-139.7(3.7)$  & 45            & $-466.7(0.7)$    & 8100         \\  \hline	
	\end{tabular}
	\caption{Performance of DRN and other regression methods for all datasets. $L2$ denotes the $L2$ loss measurement following \citet{oliva2014fast}, $NLL$ denotes the negative log-likelihood and $N_{param}$ is the number of model parameters used. For $L2$ and $NLL$, lower values reflect better regression accuracies. The number in the parentheses is  the standard error of the performance measures, over repeated runs with random initializations. DRN achieves the best performance for each dataset while using the least number of parameters.}
	\label{table:allres}
\end{table}

\subsection{Ornstein-Uhlenbeck process}
\begin{figure}[]
	\centering
	\includegraphics[width=1.0\linewidth]{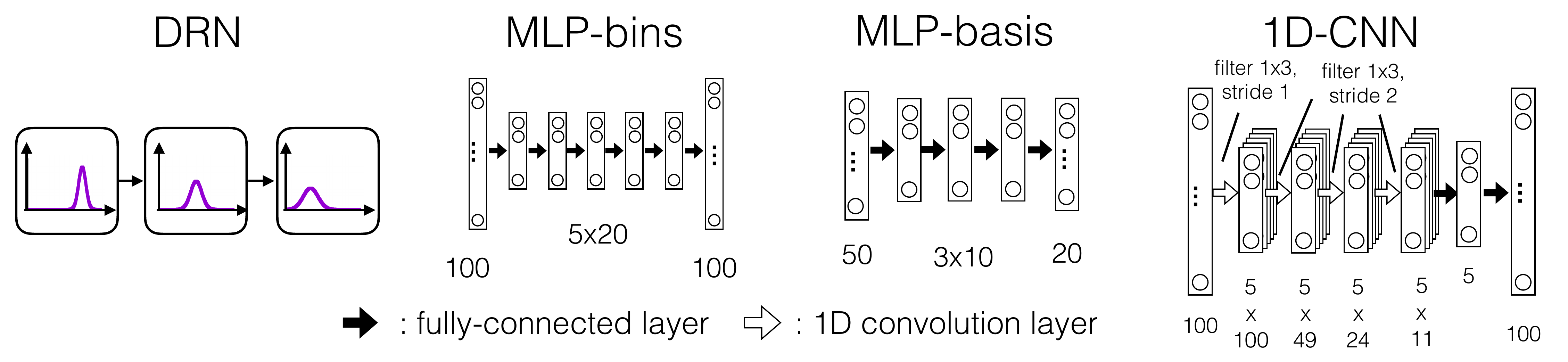}
	\caption{A comparison of the DRN, MLP (on discretized bins), MLP (on sinusoidal basis coefficients) and 1D-CNN network architectures for the OU process experiment. The DRN network consists of one hidden node in between the input and output nodes, and has only 2 weight and 8 bias parameters. For MLP-bins, MLP-basis and 1D-CNN, the numbers below each layer denotes the number of nodes. In 1D-CNN, the convolutional filters have a width of 3 and 5 feature maps.}
	\label{fig:OU_network_compare}
\end{figure}

Because of the Boltzmann distribution term (ref. Eq. (\ref{eq:condprob})), DRN models the diffusion process very well. For this experiment, we evaluate our model on data generated from the stochastic Ornstein-Uhlenbeck (OU) process which combines the notion of random walk with a drift towards a long-term mean \citep{uhlenbeck1930theory}. The OU process has wide-ranging applications. In the commodity market, prices exhibit mean-reverting patterns due to market forces and hence modelling the prices with the OU process helps form valuation strategies \citep{schwartz2000short,zhang2012efficient}. The OU process is also used to model the stochastic resonance of neurons in the brain \citep{plesser1996stochastic}.

The OU process is described by a time-varying Gaussian distribution. With the long-term mean set at zero, the pdf has a mean of $\mu(t) = y\exp(-\theta t)$ and variance of $\sigma^2(t) = \frac{D(1-e^{-2\theta t})}{\theta}$. $t$ represents time, $y$ is the initial point mass position, and $D$ and $\theta$ are the diffusion and drift coefficients respectively. The regression task is to map from an initial Gaussian distribution at $t_{init}$ to the resulting distribution after some time step $\Delta t$. The Gaussian distributions are truncated with support of $[0,1]$. With different sampled values for $y \in [0.3,0.9]$ and $t_{init} \in [0.01,2]$, pairs of distributions are created for $\Delta t = 1$, $D = 0.003$ and $\theta = 0.1$.

All methods were trained with 100 training data; we also tried increasing the training size  and attained similar results. Following \citet{oliva2014fast}, the regression performance is measured by the $L2$ loss. The results are shown in Table \ref{table:allres} and the network architectures for DRN, MLP and 1D-CCN are shown in Fig.~\ref{fig:OU_network_compare}. DRN attained the best test $L2$ loss with a small network containing a single hidden layer of one node, with a total of 2 weights and 8 bias parameters. In contrast, MLP-bins, MLP-basis, 1D-CNN and 3BE require more parameters but achieved lower regression accuracies. DRN achieves a compact network because the diffusion and drift effects of OU process are represented by the propagation properties of peak spreading and peak shifting respectively, as illustrated in Fig. \ref{fig:propbehave}.  Using a few network parameters, DRN is able to represent the physical behavior of the OU process and achieve better test accuracies.

\subsection{Brownian motion with Fokker-Planck equation}
Noise plays an important role in all physical systems and the fundamental equation to describe evolving systems influenced by noise is the Fokker-Planck equation \citep{risken1996fokker}. The Fokker-Planck equation is a differential equation for Markovian stochastic processes, relating how a probability distribution changes under drift and diffusion forces. Due to space constraints, we refer the reader to a good introductory book by \citet{risken1996fokker}. The Fokker-Planck equation has been used in a wide range of applications such as  plasma physics \citep{peeters2008fokker},  structural safety \citep{tsurui1986application}, chemistry \citep{gillespie2002chemical}, animal swarm behavior \citep{kolpas2007coarse}, finance \citep{friedrich2000quantify} and many more \citep{edwards2013grid,cohn1978stellar,chavanis2008nonlinear,hirshman1976approximate,kim2003light,walton2000equilibrium,jafari2003stochastic,mccaskill1979fokker}. 

\begin{figure*}[h]
	\centering
	\begin{subfigure}[b]{0.35\textwidth}
		\includegraphics[width=\textwidth]{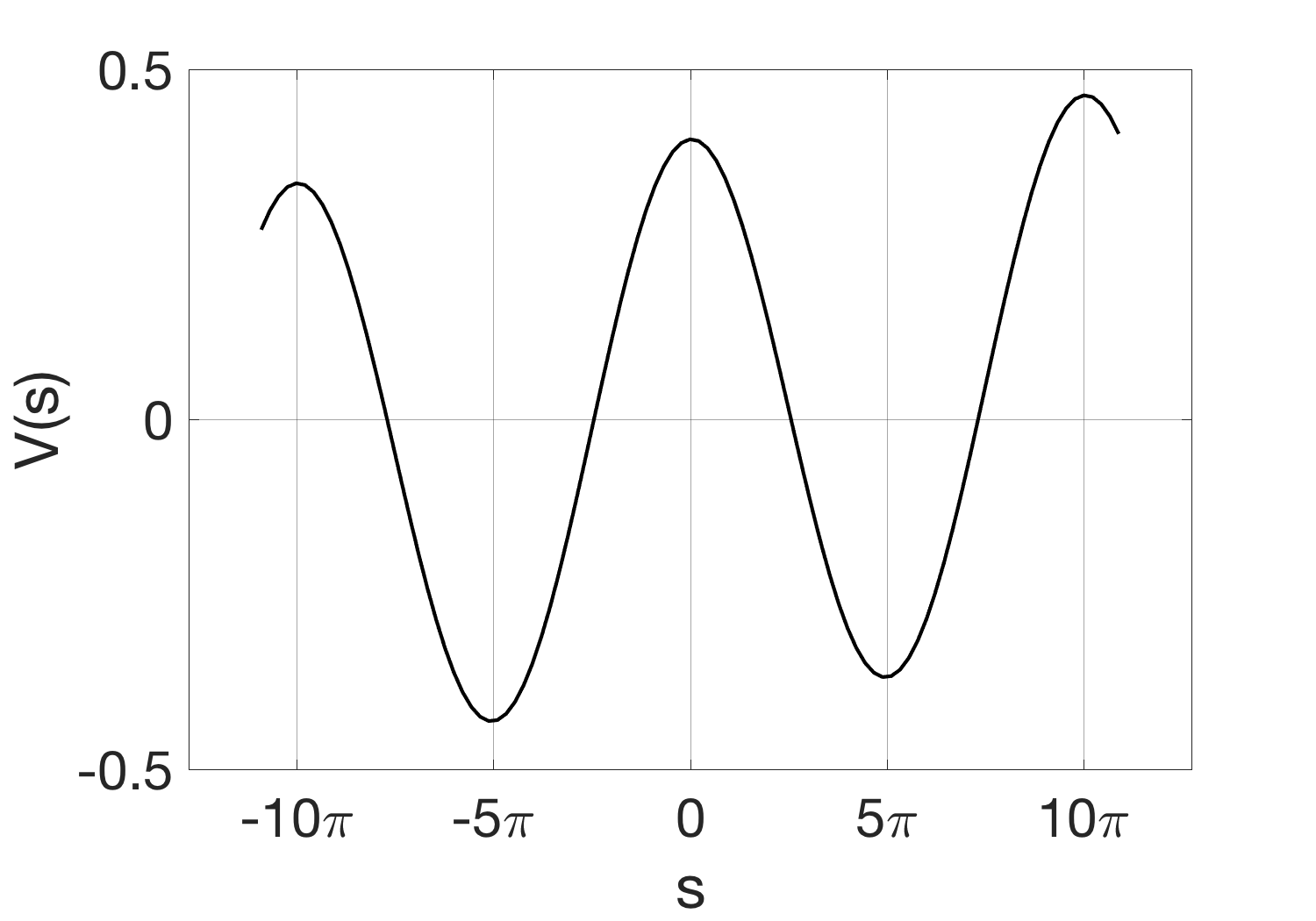}
		\caption{ }
		\label{fig:potential}
	\end{subfigure}
	\quad
	\begin{subfigure}[b]{0.55\textwidth}
		\includegraphics[width=\textwidth]{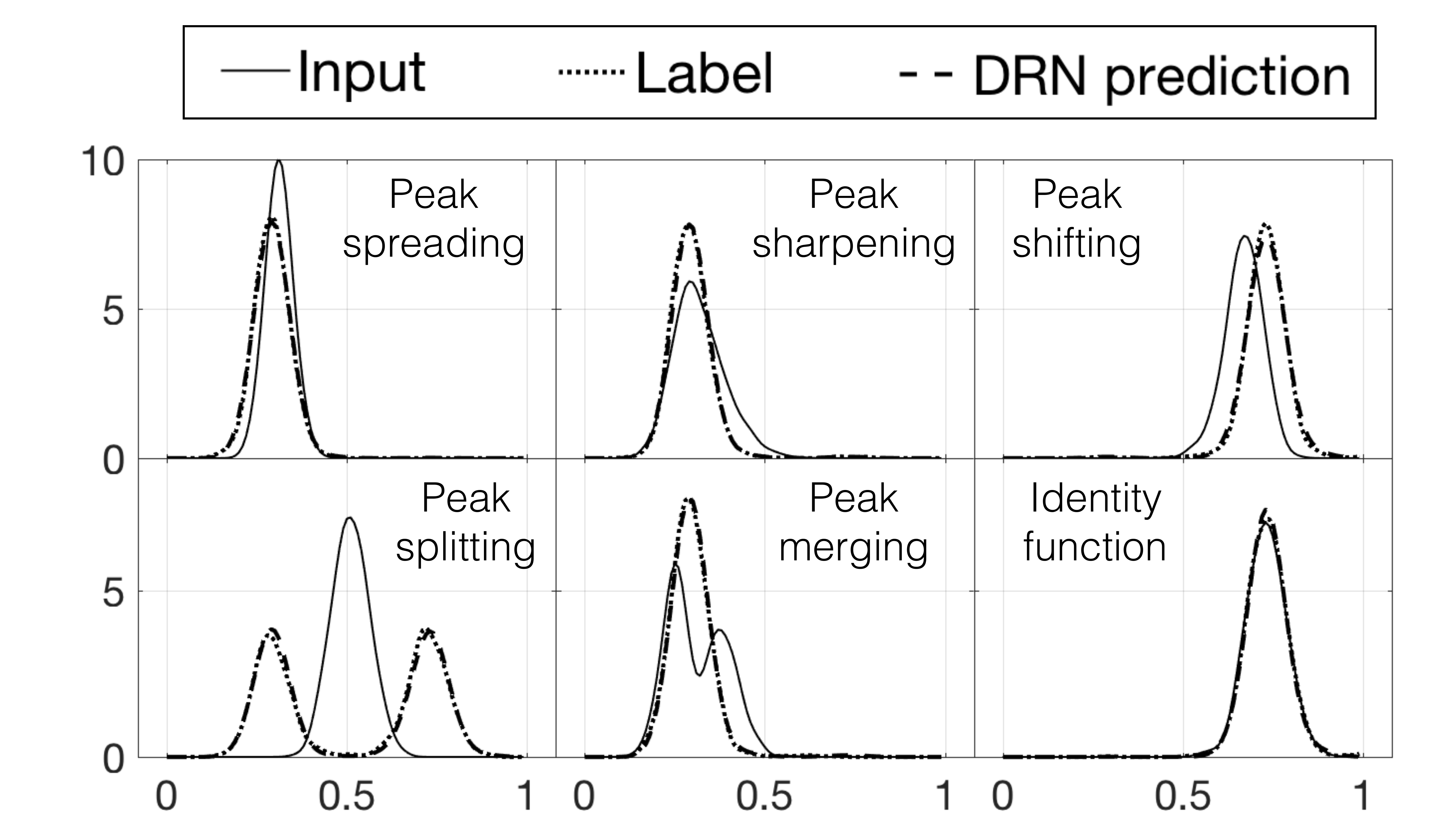}
		\caption{ }
		\label{fig:FPperiod_5egs}
	\end{subfigure}
	\caption{(a) For the Fokker-Planck experiment, the total potential $V(s)$ acting on the particles is a combination of a sinusoidal potential and a constant external force. (b) Example input-output pairs from the test data, with characteristics such as peak spreading, peak sharpening, peak shifting, peak splitting and peak merging and the identity function. DRN's prediction for each data closely matches the ground truth.}
	\label{fig:FP}
\end{figure*}

We investigate the suitability of our Distribution Regression Network for learning time-varying distributions governed by the Fokker-Planck equation. In particular, we study the problem of Brownian motion in a periodic potential \citep{risken1996fokker} which is applicable in many fields \citep{josephson1962possible, ambegaokar1969voltage,fulde1975problem,geisel1979continuous,dieterich1977diffusion,haken1967theory,lindsey1972synchronization}. In this experiment, our system consists of a one-dimensional sinusoidal potential with a constant external force. The resulting total potential is $V(s) = -0.4\cos (0.2s) - 0.002s,$ where $s$ is the random variable with support of $[-11\pi,11\pi]$ (see Fig. \ref{fig:potential}). With this potential, we solve the Fokker-Planck equation: $
\frac{\partial}{\partial t} P(s,t) = -\frac{\partial}{\partial x} V'(s)P(s,t) + \frac{1}{2}\frac{\partial^2}{\partial s} \sigma^2 P(s,t)$, with diffusion coefficient $\sigma=3.0$, to derive the probability distribution at various time steps.\footnote{Source of solver code: www-math.bgsu.edu/$\sim$zirbel/sde/matlab/index.html} We create our regression data as follows: The random variable $s$ is discretized into 100 bins. For each data, the distribution at $t=0$ is initialized to start at either one or two positions. For single position, we set the pmf to be 1.0 at a randomly sampled bin while for two positions, we set the pmf to be 0.5 at two randomly sampled bins. The regression task is to map from an initial distribution at $t_{init} \in [1,5]$  to the resulting distribution after some fixed time step $\Delta t = 10$. 1000 training data were generated. Fig. \ref{fig:FPperiod_5egs} shows some example data pairs, with transformations such as peak spreading, peak sharpening, peak shifting, peak splitting, peak merging and the identity function. For each distribution, 1000 samples were generated. For DRN, MLP-bins and 1D-CNN, kernel density estimation was performed to estimate the pdf while MLP-basis and 3BE derive the sinusoidal basis coefficients from the samples.

The regression performance is measured by the negative log-likelihood of the test samples, following \citet{oliva2013distribution}. The performance is shown in Table \ref{table:allres} where lower negative log-likelihood is favorable. DRN achieves the best regression performance with the smallest number of model parameters. We show DRN's predicted distributions on some test data in Fig. \ref{fig:FPperiod_5egs} where the prediction closely matches the ground truth label. The distribution transformations shown in Fig. \ref{fig:FPperiod_5egs} closely relates to DRN's propagation behavior described in Fig. \ref{fig:propbehave}. This validates that DRN's formulation and propagation behavior allows it to learn distribution transformations following the Fokker-Planck equation more efficiently than traditional neural networks.

\subsection{Cell length data}
Next we conducted experiments on a real-world cell dataset similar to the one used in \citet{oliva2013distribution}. The dataset is a time-series of images of fibroblast cells (see  \ref{sect:appen_exptdetails} for sample images). At each time-frame, given the distribution of cell length, we predict the distribution of the cell width. The regression performance is measured by the negative log-likelihood of the test samples, as in \citet{oliva2013distribution}. The performance is shown in Table \ref{table:allres} where lower negative log-likelihood is favorable. DRN has the lowest negative log-likelihood, using the simplest network of one input node connecting to one output node. In contrast, the other methods use more parameters but achieve poorer performance.

\begin{figure*}[h]
	\centering
	\begin{subfigure}[b]{0.49\textwidth}
		\includegraphics[width=\textwidth]{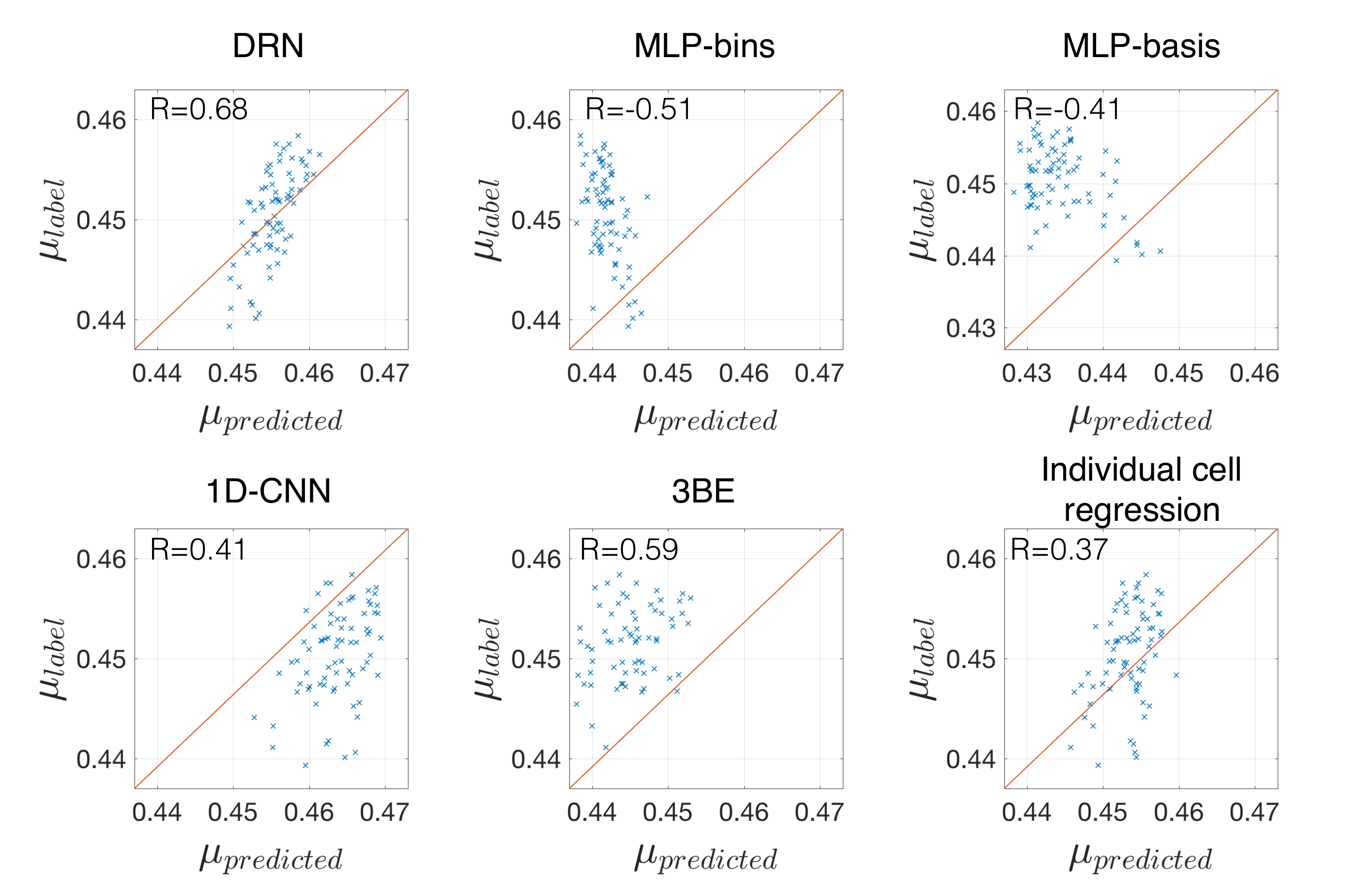}
		\caption{}
		\label{fig:cellres_mean}
	\end{subfigure}
	\begin{subfigure}[b]{0.49\textwidth}
		\includegraphics[width=\textwidth]{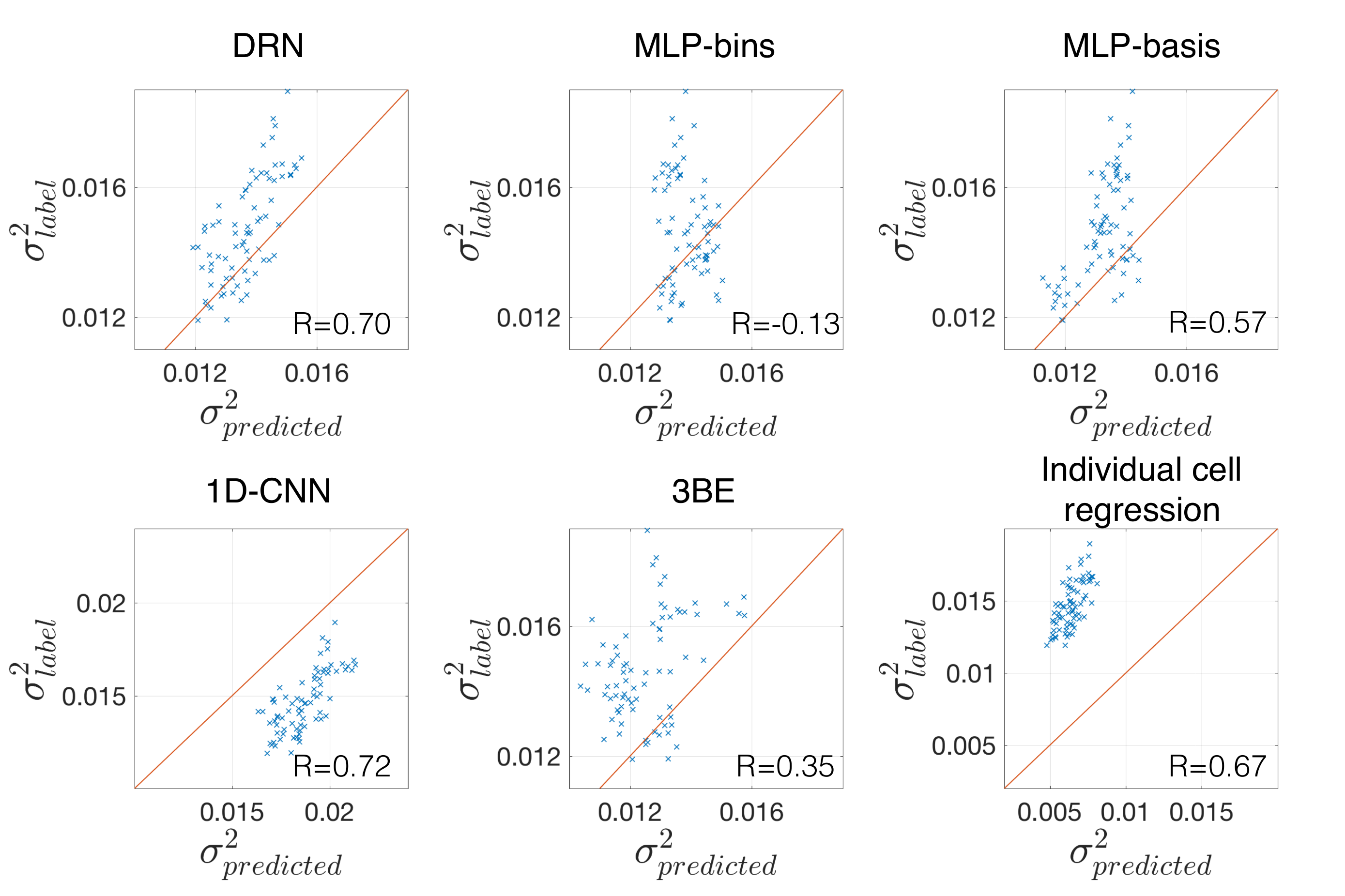}
		\caption{}
		\label{fig:cellres_var}
	\end{subfigure}
	\caption{Comparison of the (a) mean and (b) variance of the label distributions and predicted distributions on the cell length test set. DRN outperforms the rest as its data points are closest to the diagonal line and it has the highest correlation coefficients, as denoted by R).}
	\label{fig:cellres}
\end{figure*}

To visualize the regression results on the test set, we compare the first two moments (mean and variance) of the predicted and the ground truth distributions (see Fig. \ref{fig:cellres_mean} and Fig. \ref{fig:cellres_var}). Each point represents one test data and we show the correlation coefficients between the predicted and labeled moments. DRN has the best regression performance as the points lie closest to the diagonal line where the predicted and labeled moments are equal, and its correlation values are highest. 

Next, we demonstrate the need for regressing the distributions instead of treating individual cells as independent objects. In principle, one could perform regression between the length and width of individual cells and then build the cell width distribution from the regression results. However, direct regression does not capture the biological variations of cell widths (ie. cells with the same length can have different widths). We trained an MLP where the input is an individual cell's length and the output is its width. The `Individual cell regression' plots in Fig. \ref{fig:cellres} show the moments for the output distributions. While the predicted means are close to the labeled means, the variance of the output distributions are underestimated. This is analyzed further by studying the regression function learnt, which we elaborate in  \ref{sect:appen_exptdetails}. We also measure the negative log-likelihood by performing density estimation on the predicted widths where the kernel bandwidth is set following \citet{bowman1997applied}. The negative log-likelihood is -64.48, which is much poorer than the other methods, validating that regressing on individual cells does not capture the underlying stochastic noise of the data. This demonstrates that distribution-to-distribution regression is fundamentally different from performing many real-to-real regression and then aggregating the result into a distribution.

\begin{figure}[h]
	\centering
	\includegraphics[width=0.5\linewidth]{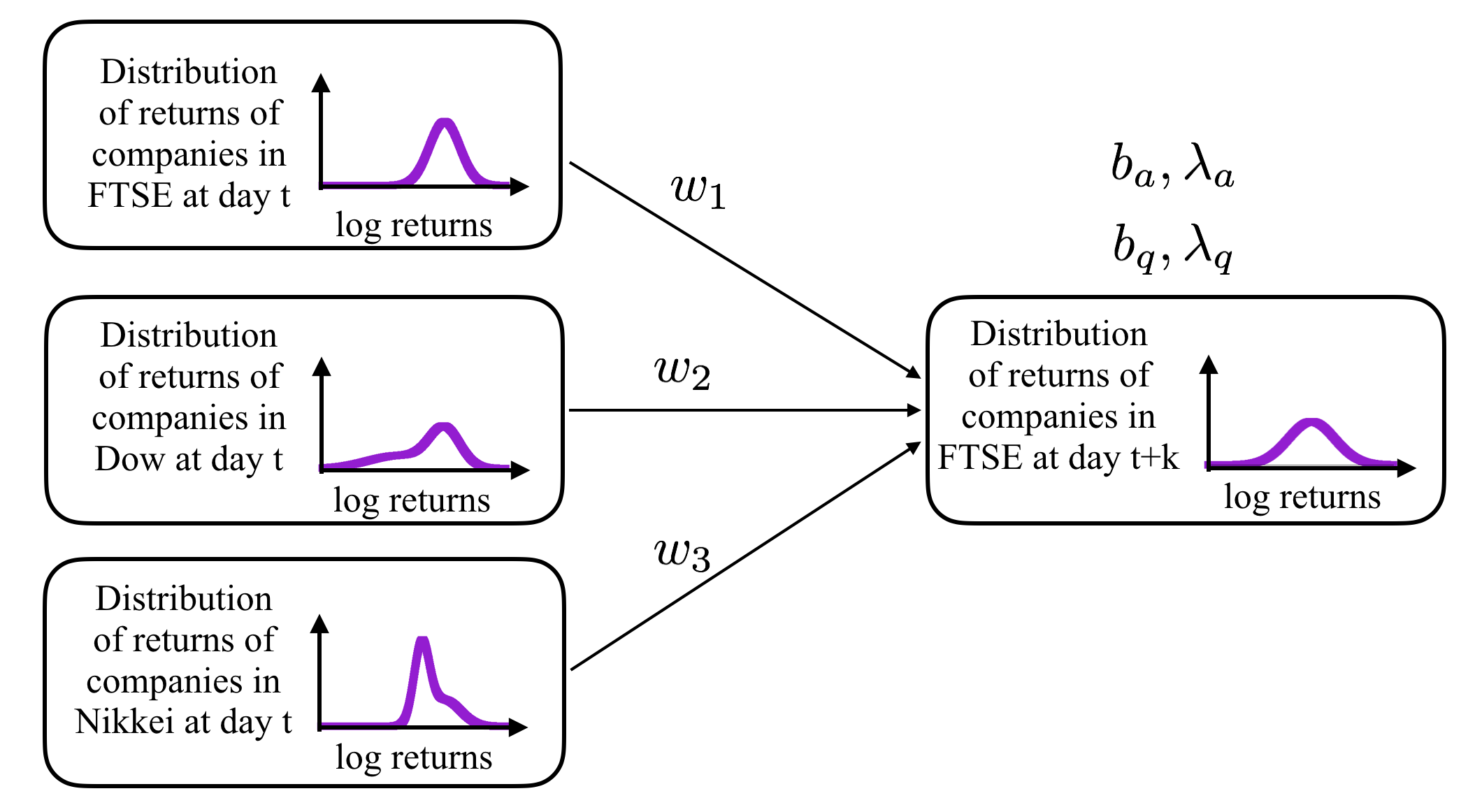}
	\caption{Single-layer network used in DRN for the stock data.}
	\label{fig:drn_stock_nw}
\end{figure}

\subsection{Stock data}
We demonstrate that DRN can be useful for a real-world stock prediction task and outperforms MLP and 3BE. With greater integration of the global stock markets, there is significant co-movement of stock indices \citep{hamao1990correlations, chong2008international}. In a study by \citet{vega2012forecasting}, it was found that the previous day stock returns of the Nikkei and Dow Jones Industrial Average (Dow) are good predictors of the FTSE return. Modelling the co-movement of global stock indices has its value as it facilitates investment decisions.

Stock indices are weighted average of the constituent companies' prices in a stock exchange, and existing research has focused on the returns of the indices. However, studies have shown improved portfolio selection when predicting the entire distribution of stock returns \citep{cenesizoglu2008distribution}. Our regression task is as follows: given the current day's distribution of returns of constituent companies in FTSE, Dow and Nikkei,  predict the distribution of returns for constituent companies in FTSE $k$ days later. The stock data consists of 9 years of daily returns from 2007 to 2015. To adapt to changing market conditions, we use a sliding-window training scheme  \citep{kaastra1996designing} (details are in \ref{sect:appen_exptdetails}). Following common practice \citep{murphy1999technical}, we performed exponential window averaging on the price series of each stock over 50 days to reduce noise.

\begin{figure}[h!]
	\centering
	\includegraphics[width=0.9\linewidth]{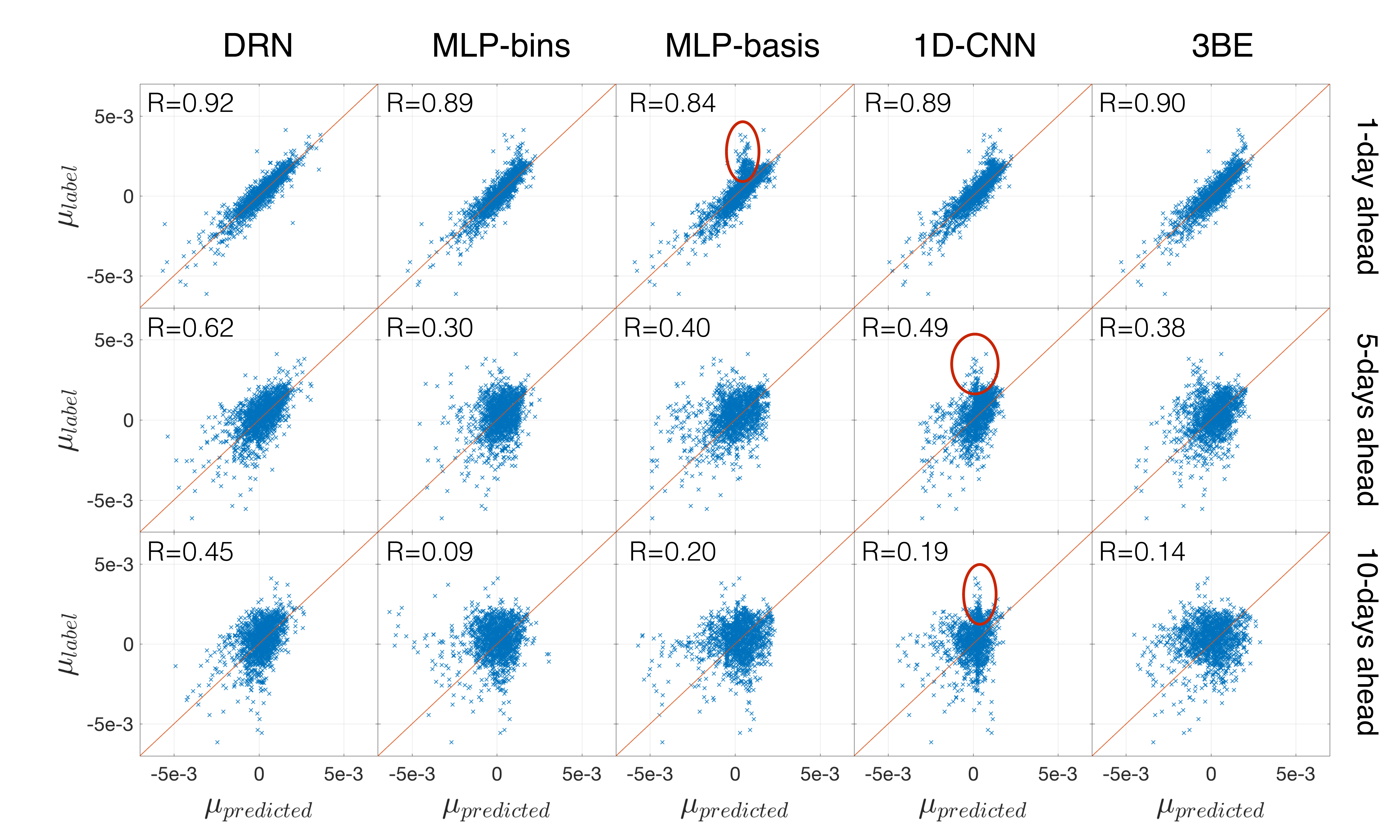}
	\caption{Comparison of the mean of the label and predicted distributions on the stock data, for varying number of days-ahead. DRN outperforms the rest as its data points are closest to the diagonal line. The other methods exhibit regression to the mean, as marked by the red ovals.}
	\label{fig:stockres_mean}
\end{figure}
\begin{figure}[h!]
	\centering
	\includegraphics[width=0.9\linewidth]{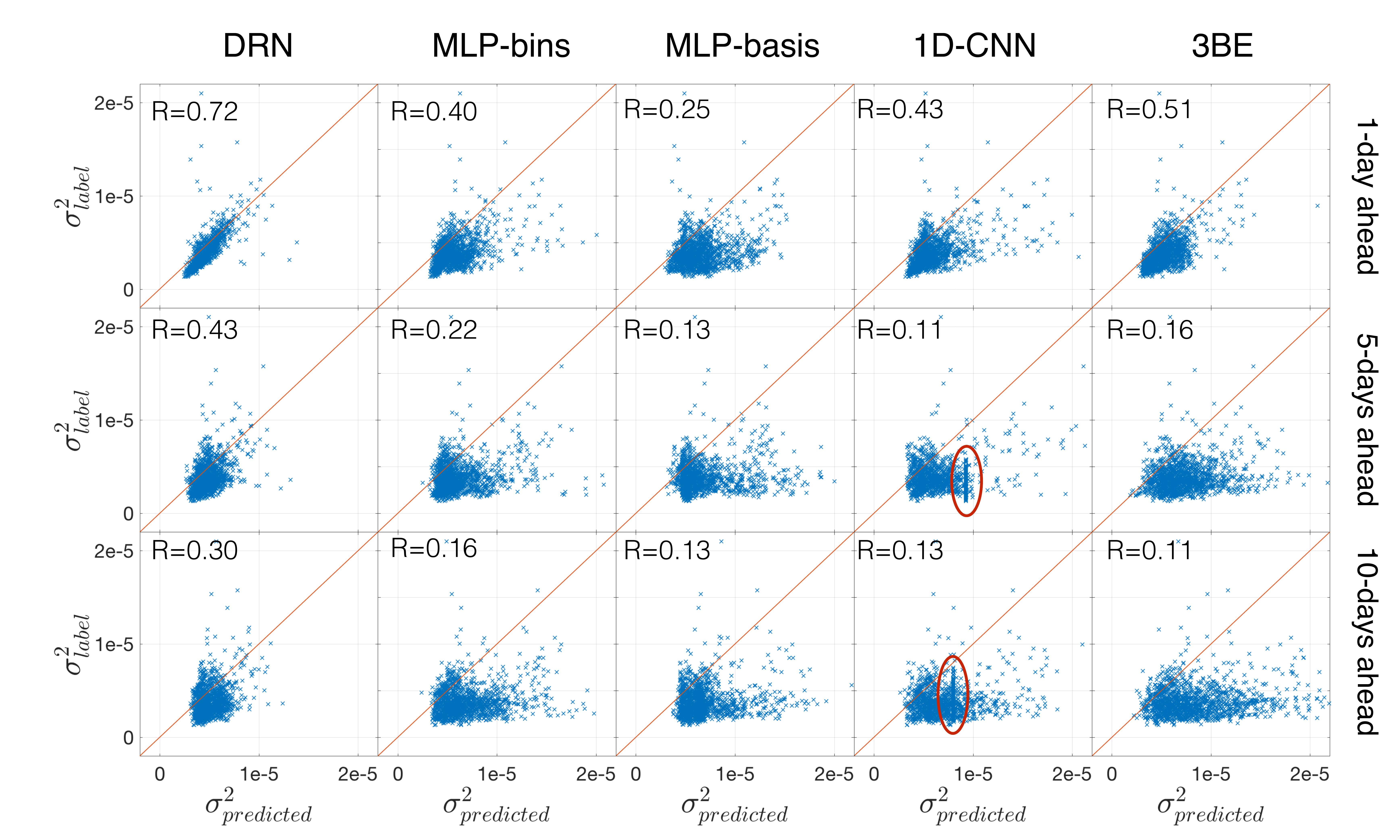}
	\caption{Comparison of the variance  of the label and predicted distributions on the stock data. DRN outperforms the rest as its data points are closest to the diagonal line. 1D-CNN outputs show some anomaly where the predictions of variance for several data are the same, as marked by the red ovals.}
	\label{fig:stockres_var}
\end{figure}

\begin{figure}[h!]
	\centering
	\includegraphics[width=0.7\linewidth]{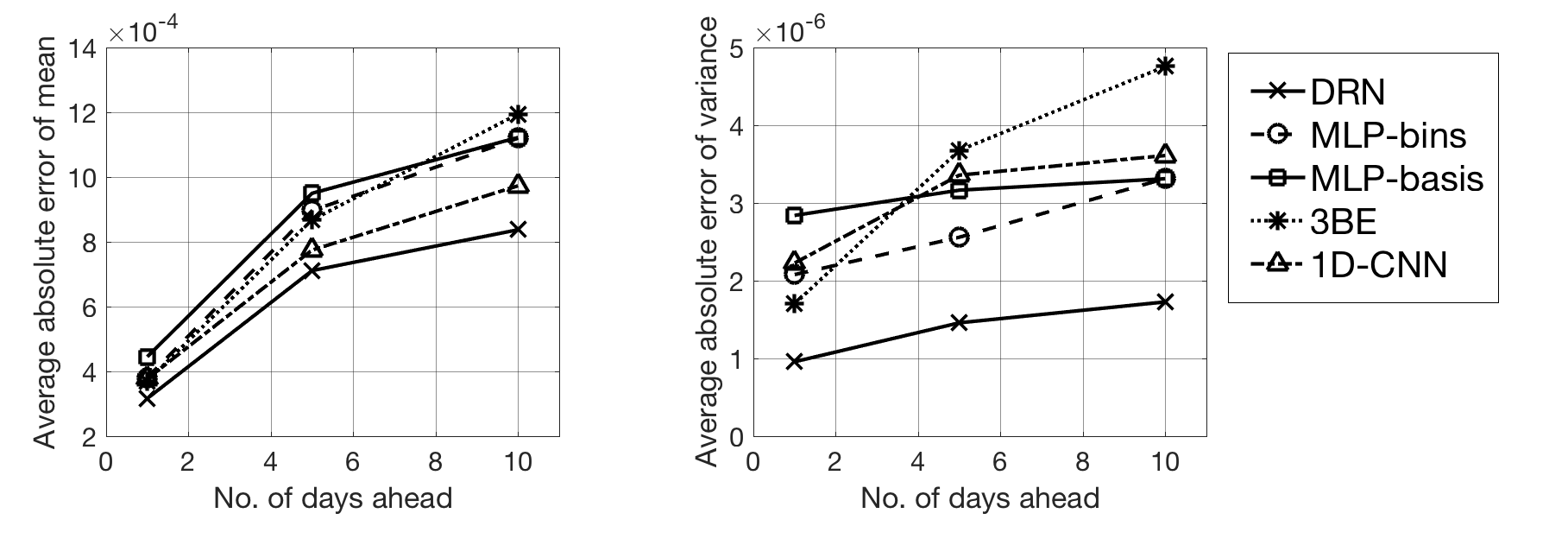}
	\caption{The average absolute error of mean and variance with predicting several days-ahead, where DRN's error is consistently the lowest.}
	\label{fig:meanvardiff_stock}
\end{figure}

We performed predictions on the next-day distributions. Table \ref{table:allres} shows the result of the negative log-likelihood. The single-layer network in DRN (see Fig. \ref{fig:drn_stock_nw}) performs well, using only 7 network parameters. In comparison, the other methods require at least 300 times more parameters. The moment plots are shown in Fig. \ref{fig:stockres_mean} and \ref{fig:stockres_var}. DRN has the best regression performance as the points lie closest to the diagonal line and its correlation values are highest. As an extension, we predict the FTSE returns distribution several days ahead. Expectedly, the performance deteriorates as the number of days increases. Still, DRN's performance remains the best, while the other methods exhibit some regression towards the mean which worsens when predicting more days ahead (see red circles in Fig. \ref{fig:stockres_mean}). In particular, 1D-CNN's variance plots show several data obtaining the same predicted variance regardless of the true variance. Fig. \ref{fig:meanvardiff_stock} summarizes the results by showing the average absolute error of the mean and variance. Even as the prediction task increases in difficulty, DRN consistently has the lowest error.

\section{Comparison of prediction times}\label{sect:compare_time}
We compare the mean prediction times in Table \ref{table:predictiontime}. MLP-basis is the fastest, followed by MLP-bins. This is expected as the computations consist of highly-optimized matrix multiplications and MLP-basis has fewer network parameters than MLP-bins. This is followed by DRN and 1D-CNN. DRN is fast due to the small number of parameters and the matrix multiplications in Eq. (\ref{eq:factorizeP}) and 1D-CNN has fewer parameters than MLP because of the tied weights. 3BE has the longest prediction time as it involves many computation steps. Although runtime is important, we first need a compact model that represents the distributions well. Now that we have a compact DRN model, we can work on computation speed up in future work.
\begin{table}[h!]
	\centering
	\begin{tabular}{ccccc}
		\toprule
		& \multicolumn{4}{c}{Mean prediction time (ms)}                                                                                                                                  \\ \cmidrule{2-5} 
		& \begin{tabular}[c]{@{}c@{}}OU\end{tabular} &
		\begin{tabular}[c]{@{}c@{}}Fokker-Planck\end{tabular} &
		 \begin{tabular}[c]{@{}c@{}}Cell\end{tabular} & \begin{tabular}[c]{@{}c@{}}Stock \end{tabular}  \\ \midrule
		\multicolumn{1}{c}{DRN}    & 0.01  &  0.34  & 0.01     &0.03  \\ 
		\multicolumn{1}{c}{MLP-bins} & 0.01 & 0.01  &0.06    & 0.01       \\ 
		\multicolumn{1}{c}{MLP-basis} & 0.005&  0.003 &0.005    & 0.005      \\ 
		\multicolumn{1}{c}{1D-CNN}   & 0.03   & 0.02 & 0.20  & 0.08       \\ 
		\multicolumn{1}{c}{3BE}  & 0.98  & 1.25 & 0.32   & 5.95        \\ 
		\bottomrule
	\end{tabular}
	\caption{Mean prediction time per data, all runs were conducted on the CPU. }
	\label{table:predictiontime}
\end{table}
\section{Discussion}\label{sect:conclude}
The encoding of functions in neural networks poses a challenge in the task of function-to-function regression. Since each node in the network encodes only a real value, the function has to be broken into smaller parts, either via discretized grid of points or basis functions, and encoded using multiple nodes. This results in larger number of parameters which may give rise to several problems such as difficulty in optimization, overfitting and large memory usage. To that end, we propose a novel idea of encoding an entire function in a single node. We present such a model for the distribution regression task. Our Distribution Regression Network (DRN) is a compact network model for distribution regression that exhibits useful propagation behavior. We benchmark DRN's performance with the multilayer perceptron, one-dimensional convolutional neural network and the state-of-the-art Triple-Basis Estimator. For the datasets tested, DRN achieved higher test accuracies while using fewer parameters. In particular, on the stock prediction dataset, DRN uses at least 300 times fewer parameters than the rest. This validates that by encoding entire functions within a single node, DRN is able to achieve compact networks for the distribution regression task.

Our idea of encoding a function in a single node can be extended to other applications. For future work, we look to extend DRN for variants of the distribution regression task such as distribution-to-real regression \citep{oliva2014fast} and distribution classification. Another possibility is to adapt DRN for handling sequences of data using architectures similar to recurrent neural networks. Extensions may also be made for handling multivariate distributions.

\section{Acknowledgements}
We thank Kuan Pern Tan and Jorge Sanz for their constructive discussions. This work was supported by the Biomedical Research Council of A*STAR (Agency for Science, Technology and Research), Singapore and the National University of Singapore.

\appendix
\section{Derivations of cost gradients} \label{sect:appen_costgrad}
In Section 2.4, we presented the key equation for deriving backpropagation gradients:
\begin{align}
\frac{\partial P^{(L)}_1 \left(s^{(L)}_1\right)}{\partial P^{(l)}_k \left(s^{(l)}_k\right)} = \sum_j^n \sum_{s^{(l+1)}_j} \frac{\partial P^{(L)}_1 \left(s^{(L)}_1\right)}{\partial P^{(l+1)}_j \left(s^{(l+1)}_j\right)} \frac{\partial P^{(l+1)}_j \left(s^{(l+1)}_j\right)}{\partial P^{(l)}_k \left(s^{(l)}_k\right)}
\label{eq:append_backprop}
\end{align}
To derive the cost gradients for optimization, we need the derivative of the final output node distribution with respect to the network parameters. For instance, for the network weights:
\begin{align}
\frac{\partial P^{(L)}_1 \left(s^{(L)}_1\right)}{\partial w_{ki}^{(l)}} =  \sum_{s^{(l)}_k}
\frac{\partial P^{(L)}_1 \left(s^{(L)}_1\right)}{\partial P^{(l)}_k \left(s^{(l)}_k\right)} 
\frac{\partial P^{(l)}_k \left(s^{(l)}_k\right)} {\partial w_{ki}^{(l)}}
\end{align}
We first derive the intermediate gradient terms required to obtain the gradients in Eq. (\ref{eq:append_backprop}) and proceed to derive gradients for the network parameters. All of the equations work on the discretized distributions, where the integrals are now expressed in summations.
\subsection{Derivation of intermediate backpropagation terms}
We start with deriving the final term of Eq. (\ref{eq:append_backprop}) which is the gradient of the upper layer node distribution with respect to the incoming lower node distribution. The subscripts and superscripts are renamed for ease of explanation in later derivations.
\begin{align}
\frac{\partial P^{(l)}_k \left(s^{(l)}_k\right)}{\partial P^{(l-1)}_i \left(s^{(l-1)}_i\right)} 
= \sum_{s^{(l)'}_k} \frac{\partial P^{(l)}_k \left(s^{(l)}_k\right)}{\partial \tilde{P}^{(l)}_k \left(s^{(l)'}_k\right)} \frac{\partial \tilde{P}^{(l)}_k \left(s^{(l)'}_k\right)}{\partial P^{(l-1)}_i \left(s^{(l-1)}_i\right)} 
\label{eq:append_gradnodenode}
\end{align}
The above derivative is a consequence of the normalization step taken in Eq. (\ref{eq:norm}) of the main manuscript. At each node, we need to compute the derivative of the normalized distribution with respect to the unnormalized distribution. Recall $P^{(l)}_k \left(s^{(l)}_k\right) = \frac{ \tilde{P}^{(l)}_k \left(s^{(l)}_k\right)}{Z^{(l)}_k },$ where $Z^{(l)}_k = \sum_{s^{(l)'}_k} \tilde{P}^{(l)}_k \left(s^{(l)'}_k\right)$.
\begin{align}
\frac{\partial P^{(l)}_k \left(s^{(l)}_k\right)}{\partial \tilde{P}^{(l)}_k \left(s^{(l)'}_k\right)}
=     \begin{cases}
\frac{1}{Z^{(l)}_k} - \frac{\tilde{P}^{(l)}_k \left(s^{(l)}_k\right)}{\left(Z^{(l)}_k\right)^2}       & \quad \text{if } s^{(l)}_k = s^{(l)'}_k\\
-\frac{\tilde{P}^{(l)}_k \left(s^{(l)}_k\right)}{\left(Z^{(l)}_k\right)^2}  & \quad \text{otherwise }
\end{cases}
\end{align}
For the final term of Eq. (\ref{eq:append_gradnodenode}), $\frac{\partial \tilde{P}^{(l)}_k \left(s^{(l)'}_k\right)}{\partial P^{(l-1)}_i \left(s^{(l-1)}_i\right)} $, the derivation proceeds from the propagation step in Eq.~(\ref{eq:factorizeP}), reproduced here in discrete form.
\begin{align}
\label{eq:append_factorizeP}
\tilde{P}_k^{(l)} \left(s_k^{(l)}\right) &=  \exp\left(B\left(s_k^{(l)}\right) \right)  \prod_i^n \left\{ \sum_{s_i^{(l-1)}} P_i^{(l-1)}\left(s_i^{(l-1)}\right) \exp\left[ - w_{ki}^{(l)}
\left( \frac{s_k^{(l)}-s_i^{(l-1)}}{\Delta}\right)^2  \right]  \right\} 
\end{align}
By substituting 
\[\Gamma_i \left(s_k^{(l)}\right) = \sum_{s_i^{(l-1)}} P_i^{(l-1)}\left(s_i^{(l-1)}\right) \exp\left[ - w_{ki}^{(l)}
\left( \frac{s_k^{(l)}-s_i^{(l-1)}}{\Delta}\right)^2  \right],  \]
we obtain
\begin{align}
\label{eq:append_factorizeP_short}
\tilde{P}_k^{(l)} \left(s_k^{(l)}\right) =  \exp\left(B\left(s_k^{(l)}\right) \right)  \prod_i^n  \Gamma_i \left(s_k^{(l)}\right).
\end{align}
Eq. (\ref{eq:append_factorizeP_short}) is a product of variables and its derivative with respect to any variable is obtained by product rule.
\begin{align}
\frac{\partial \tilde{P}_k^{(l)} \left(s_k^{(l)}\right)}{\partial x} 
= \tilde{P}_k^{(l)} \left(s_k^{(l)}\right) \left( \frac{\frac{\partial  \exp\left(B\left(s_k^{(l)}\right) \right)  }{\partial x}}{ \exp\left(B\left(s_k^{(l)}\right) \right)  } 
+ \sum_i^n \frac{\frac{\partial \Gamma_i \left(s_k^{(l)}\right)}{\partial x}}{\Gamma_i \left(s_k^{(l)}\right)}\right),
\label{eq:append_prodrule}
\end{align}
where $x$ can be one of the lower layer node probabilities $ P_i^{(l-1)}\left(s_i^{(l-1)}\right) $, or one of the network parameters $ ( w_{ki}^{(l)}, \ b_{a,k}^{(l)}, \ b_{q,k}^{(l)}, \ \lambda_{a,k}^{(l)}, \ \lambda_{q,k}^{(l)} )$.
Now we can derive the final term of Eq. (\ref{eq:append_gradnodenode}).
\begin{align}
\frac{\partial \tilde{P}_k^{(l)} \left(s_k^{(l)}\right)}{\partial P^{(l-1)}_i \left(s^{(l-1)}_i\right)} 
&= \tilde{P}_k^{(l)} \left(s_k^{(l)}\right)  \frac{\frac{\partial \Gamma_i \left(s_k^{(l)}\right)}{\partial P^{(l-1)}_i \left(s^{(l-1)}_i\right)}}{\Gamma_i \left(s_k^{(l)}\right)}\\ \nonumber
&= \tilde{P}_k^{(l)} \left(s_k^{(l)}\right)  \frac{\exp \left[ - w_{ki}^{(l)}
	\left( \frac{s_k^{(l)}-s_i^{(l-1)}}{\Delta}\right)^2  \right]}{\Gamma_i \left(s_k^{(l)}\right)}
\end{align}
\subsection{Derivation of gradients with respect to network parameters}
The derivatives of the unnormalized probability distribution of a node with respect to the connecting weights and bias parameters can be derived from Eq. (\ref{eq:append_prodrule}). 

First, for each node, we compute the derivative of its unnormalized distribution with respect to an incoming weight.
\begin{align}
\frac{\partial \tilde{P}_k^{(l)} \left(s_k^{(l)}\right)}{\partial w_{ki}^{(l)}} 
= \frac{\tilde{P}_k^{(l)} }{\Gamma_i \left(s_k^{(l)}\right)} \frac{\partial \Gamma_i \left(s_k^{(l)}\right)}{\partial w_{ki}^{(l)}},
\end{align}
where
\begin{align}
\frac{\partial \Gamma_i \left(s_k^{(l)}\right)}{\partial w_{ki}^{(l)}}  
= \sum_{s_i^{(l-1)}} P_i^{(l-1)}\left(s_i^{(l-1)}\right) \exp\left[ - w_{ki}^{(l)}
\left( \frac{s_k^{(l)}-s_i^{(l-1)}}{\Delta}\right)^2  \right] \left[ -
\left( \frac{s_k^{(l)}-s_i^{(l-1)}}{\Delta}\right)^2 \right].
\end{align}
Similarly, for the bias parameters, we derive the gradients from Eq. (\ref{eq:append_prodrule}). Here we show for $b_{a,k}^{(l)}$,
\begin{align}
\frac{\partial \tilde{P}_k^{(l)} \left(s_k^{(l)}\right)}{\partial b_{a,k}^{(l)}} 
=\frac{ \partial \tilde{P}_k^{(l)} \left(s_k^{(l)}\right) }{ \partial \exp\left(B\left(s_k^{(l)}\right) \right)} \frac{\partial  \exp\left(B\left(s_k^{(l)}\right) \right) }{\partial b_{a,k}^{(l)}},
\end{align}
where
\begin{align}
\frac{\partial \exp\left(B\left(s_k^{(l)}\right) \right) }{\partial b_{a,k}^{(l)}}
= \frac{\partial B\left(s_k^{(l)}\right)}{\partial b_{a,k}^{(l)}}  \exp\left(B\left(s_k^{(l)}\right) \right),
\end{align}
and
\begin{align}
\frac{\partial B\left(s_k^{(l)}\right)}{\partial b_{a,k}^{(l)}}  
=  \begin{cases}
-\frac{s_k^{(l)}-\lambda_{a,k}^{(l)}}{\Delta}     & \quad \text{if } s_k^{(l)} > \lambda_{a,k}^{(l)}\\
\frac{s_k^{(l)}-\lambda_{a,k}^{(l)}}{\Delta} & \quad \text{otherwise }
\end{cases}
\end{align}
The derivatives for the other bias parameters can be obtained similarly.
\begin{align}
\frac{\partial B\left(s_k^{(l)}\right)}{\partial b_{q,k}^{(l)}}  
= - \left( \frac{s_k^{(l)}-\lambda_{q,k}^{(l)}}{\Delta} \right)^2
\end{align}
\begin{align}
\frac{\partial B\left(s_k^{(l)}\right)}{\partial \lambda_{a,k}^{(l)}}  
=  \begin{cases}
b_{a,k}^{(l)}    & \quad \text{if } s_k^{(l)} > \lambda_{a,k}^{(l)}\\
-b_{a,k}^{(l)}  & \quad \text{otherwise }
\end{cases}
\end{align}
\begin{align}
\frac{\partial B\left(s_k^{(l)}\right)}{\partial \lambda_{q,k}^{(l)}}  
&= -2  b_{q,k}^{(l)} \left( \frac{s_k^{(l)}-\lambda_{q,k}^{(l)}}{\Delta} \right) \left(-\frac{1}{\Delta} \right) \\ \nonumber
&= \frac{2 b_{q,k}^{(l)} }{\Delta^2} \left( s_k^{(l)}-\lambda_{q,k}^{(l)}\right)
\end{align}

\section{Experimental details}\label{sect:appen_exptdetails}
\subsection{Ornstein-Uhlenbeck process}
For DRN, MLP-bins and 1D-CNN, $q=100$ was used for discretization of the pdf while 10,000 samples were taken for each distribution to train 3BE. For MLP-basis, there are 50 basis functions for the input distribution and 20 basis functions for the output distribution. These values are chosen by cross-validation to optimize the regression performance. Table \ref{table:OUres} shows the details of the models.

\begin{table*}[h!]
	\centering
	\caption{Comparison of $L2$ loss on the Ornstein-Uhlenbeck data and the number of model parameters, with descriptions of the models. For DRN, MLP-bins and MLP-basis, the network
architecture is denoted as such: Eg. 1 - 4x3 - 1: 1 input node, followed by 4 fully-connected layers each having 3 nodes, and 1 output node. For 1D-CNN, all convolutional layers are one-dimensional with a filter width of 3. convf5s2 means a convolutional layer with 5 feature maps and stride 2.}
	\begin{tabular}{cccc}
		\toprule
		& $L2(\times 10^{-2})$ & Model description & No. of parameters \\  \midrule
		\multicolumn{1}{c}{DRN} &  $\mathbf{3.8(0.1)}$ & 1 - 1x1 -1  &  10                      \\ 
		\multicolumn{1}{c}{MLP-bins} & 5.2(0.2)      & 100 - 5x20 - 100 &    5800               \\ 
		\multicolumn{1}{c}{MLP-basis} & 6.0(0.3)   & 50 - 3x10 - 20 &    900               \\ 
		\multicolumn{1}{c}{1D-CNN} & 4.4(0.2)  &  100 - convf5s1 - 3 x convf5s2 - 5 - 100 &  1140       \\ 
		\multicolumn{1}{c}{3BE} &  4.5(0.3)   &  11 basis functions, 800  RKS features &  8800        \\ 
		\bottomrule
	\end{tabular}
	\label{table:OUres}
\end{table*}

\subsection{Brownian motion using Fokker-Planck equation}
The total potential consisting of a sinusoidal potential and a constant external force is $V(s) = -0.4\cos (0.2s) - 0.002s$, where $s$ is the random variable with support of $[-11\pi,11\pi]$. With this potential, we solve for the Fokker-Planck equation using a solver available online\footnote{Source of solver: www-math.bgsu.edu/$\sim$zirbel/sde/matlab/index.html}. The diffusion coefficient is set as 3.0. 1000 training data was generated and for each distribution, 1000 samples were generated from the pdf. For DRN, MLP-bins and 1D-CNN, the pdf is estimated using kernel density estimation with a Gaussian kernel of bandwidth 0.02 and discretization of q=100. Table \ref{table:FPres} shows the details of the models.

\begin{table*}[h!]
	\centering
	\caption{Comparison of negative log-likelihood ($NLL$) on the Fokker-Planck data and the number of model parameters, with descriptions of the models. For DRN, MLP-bins and MLP-basis, the network
		architecture is denoted as such: Eg. 1 - 4x3 - 1: 1 input node, followed by 4 fully-connected layers each having 3 nodes, and 1 output node. For 1D-CNN, all convolutional layers are one-dimensional with a filter width of 3. convf5s2 means a convolutional layer with 5 feature maps and stride 2.}
	\begin{tabular}{cccc}
		\toprule
		& $NLL$ & Model description & No. of parameters \\  \midrule
		\multicolumn{1}{c}{DRN} &  $\mathbf{-1291.8(0.2)}$ & 1 - 5x6 -1  &  280                     \\ 
		\multicolumn{1}{c}{MLP-bins} & -1290.7(0.5)     & 100 - 3x20 - 100 &    4960            \\ 
		\multicolumn{1}{c}{MLP-basis} & -1291.6(1.3)  & 27 - 2x5 - 22 &    302               \\ 
		\multicolumn{1}{c}{1D-CNN} & -1291.5(0.2)  &  100 - 2 x convf5s2 - 20 - 100 &  4620       \\ 
		\multicolumn{1}{c}{3BE} &  -1288.3(3.2)  &  19 basis functions,  5000 RKS features &  95000        \\ 
		\bottomrule
	\end{tabular}
	\label{table:FPres}
\end{table*}

\begin{figure*}[h!]
	\centering
	\begin{subfigure}[b]{0.34\textwidth}
		\includegraphics[width=\textwidth]{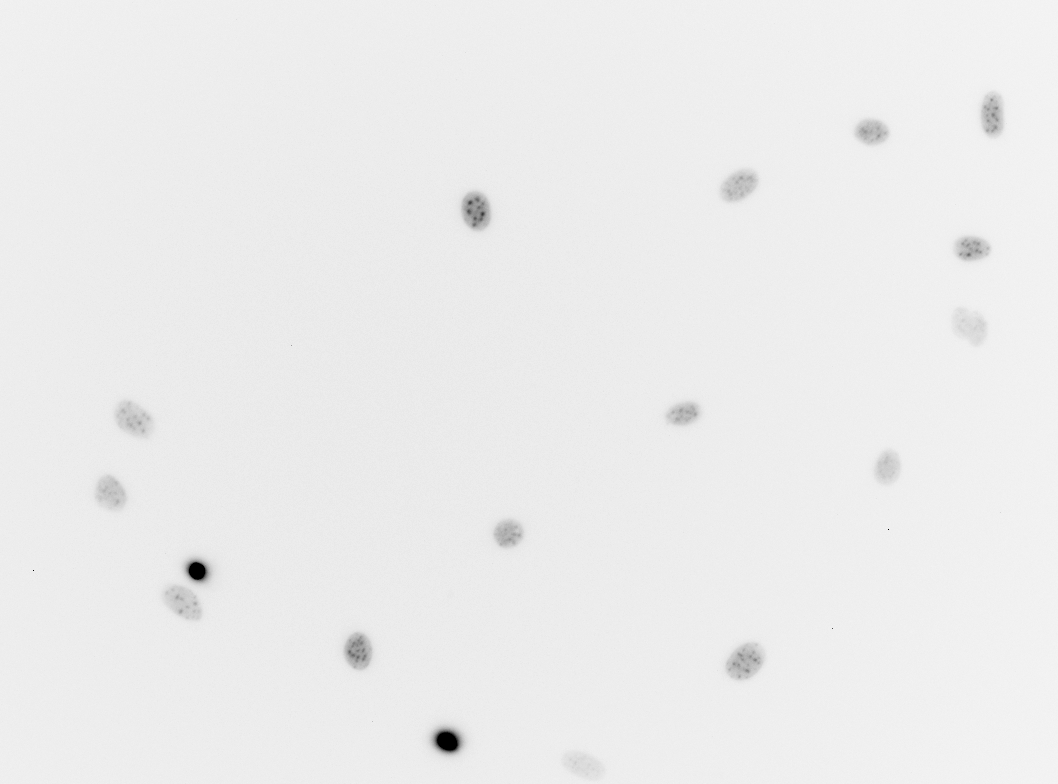}
		\caption{}
		\label{fig:sample_image}
	\end{subfigure}
	\qquad 	\qquad
	\begin{subfigure}[b]{0.45\textwidth}
		\includegraphics[width=\textwidth]{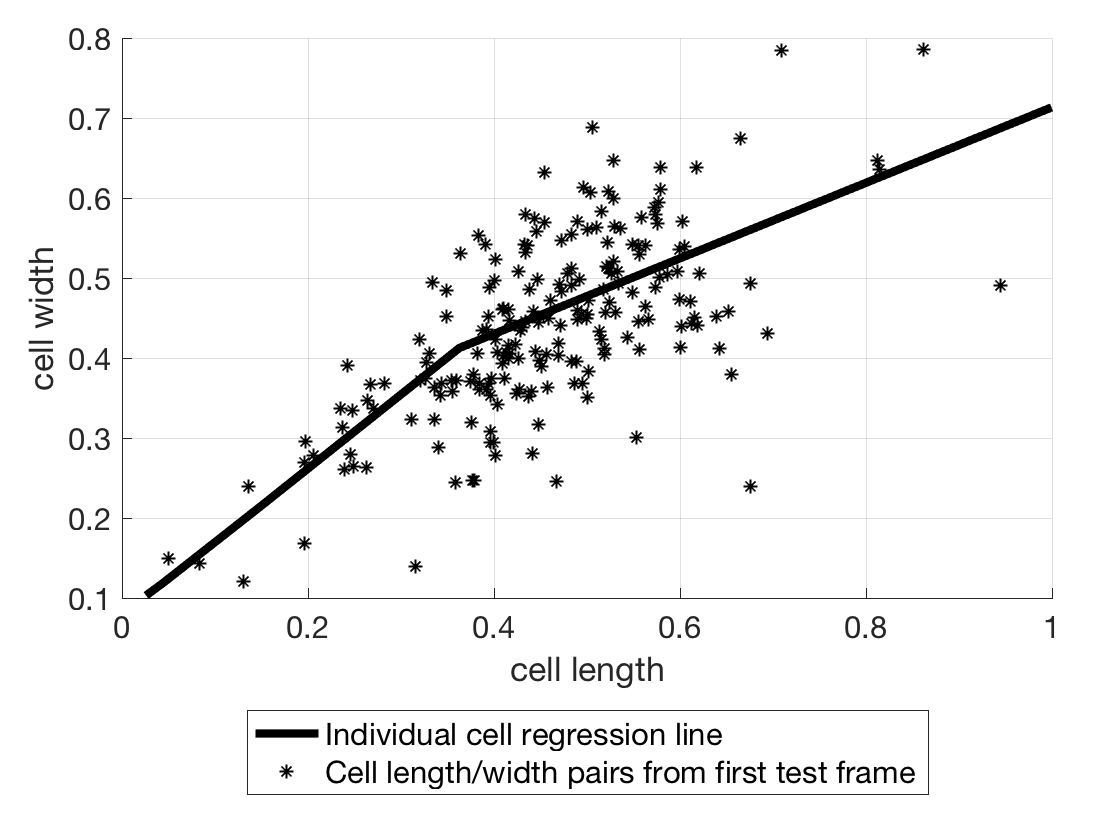}
		\caption{}
		\label{fig:indivcellreg_res}
	\end{subfigure}
	\caption{(a) Cropped image from the cell dataset. (b) Results from regressing from individual cell length to width. The line shows the regression learnt by the MLP while the points plots the individual cell's width against the length for the first image frame. This plot illustrates that regression on individual cell fails to capture the full variance of the actual measurements, resulting in the underestimated distribution variance as discussed in the main text.}
\end{figure*}
\subsection{Cell length data}
There are 277 frames containing 176 to 222 cells each. Fig. \ref{fig:sample_image} shows a crop of an image from one of the frames in the cell dataset. The elliptic shapes are the fibroblast cells. In each frame, we measured the length and width of the cells and scaled them to [0,~1]. From the 277 frames, the first 200 frames were used for training and last 77 for testing. For DRN, MLP-bins and 1D-CNN, the pdf is estimated using kernel density estimation with a Gaussian kernel of bandwidth 0.02 and discretization of $q = 100$. Table \ref{table:cellres} shows the details of the models.

\begin{table*}[h!]
	\centering
	\caption{Comparison of negative log-likelihood ($NLL$) on the cell length data and the number of model parameters, with descriptions of the models. For DRN, MLP-bins and MLP-basis, the network architecture is denoted as such: Eg. 1 - 4x3 - 1: 1 input node, followed by 4 fully-connected layers each having 3 nodes, and 1 output node. For 1D-CNN, all convolutional layers are one-dimensional with a filter width of 3. convf5s2 means a convolutional layer with 5 feature maps and stride 2. For the individual cell regression, the single input node is for the cell length while the single output node is for the cell width.}
	\begin{tabular}{cccc}
		\toprule
		& $NLL$ & Model description & No. of parameters \\  \midrule
		\multicolumn{1}{c}{DRN} &  $\mathbf{-148.5(0.4)}$ & No hidden layer  &  5                  \\ 
		\multicolumn{1}{c}{MLP-bins} & -147.8(0.1)     & 100 - 1x20 - 100 &    4120            \\ 
		\multicolumn{1}{c}{MLP-basis} & -145.7(0.4)  & 11 - 2x5 - 12 &    162              \\ 
		\multicolumn{1}{c}{1D-CNN} & -146.2(0.2)  &  100 - 3 x convf5s1 - 100 &  381       \\ 
		\multicolumn{1}{c}{3BE} &  -139.7(3.7) &  9 basis functions,  5 RKS features &  45        \\ 
		\multicolumn{1}{c}{Individual cell regression} &  -64.8(2.0)  &  1 - 20 - 1 &  61        \\ 
		\bottomrule
	\end{tabular}
	\label{table:cellres}
\end{table*}

\subsubsection{Individual cell regression results}
Here we give insights as to why regression on the individual cell results in underestimation of the variance of the output width distribution, as shown in Fig. 4(b) of the main paper. Fig. \ref{fig:indivcellreg_res} shows the regression line learnt by the MLP and also the individual cell's width against the length for the first frame in the test set. As the MLP predicts only one width value for every input of the cell length, it is unable to account for the fact that in the real biological data, cells with the same length can have different widths. Hence, regression on the individual cells leads to gross underestimation of the variance of the output width distribution. This demonstrates the need for regressing the distributions instead of treating individual cells as independent objects.

\subsection{Stock data}
The return of a company's stock is measured by the logarithmic return. The log-return of a company's stock at day $t$ is given by $\ln(V_t/V_{t-1})$, where $V_t$ represents the closing price of the company's stock at day $t$. We conduct a sliding window training scheme to account for changing market conditions. A new window is created and the network is retrained after every 300 days (which is the size of test set). For each test set, the previous 500 and 100 days were used for training and validation.

For DRN, MLP-bins and 1D-CNN, the pdf is estimated using kernel density estimation with a Gaussian kernel function with bandwidth of 0.001 and $q = 100$ was used for discretization. The authors of 3BE have extended their method for multiple input functions (see joint motion prediction experiment). We followed their method and concatenated the basis coefficients obtained from the three input distributions. Similarly for MLP-basis, we concatenated the basis coefficients from the three distributions to use as data inputs to the MLP. Table \ref{table:stockres} shows the detailed model descriptions.

\begin{table*}[h!]
	\centering
	\caption{Comparison of negative log-likelihood ($NLL$) on the stock data and the number of model parameters, with descriptions of the models. For DRN, MLP-bins and MLP-basis, the network architecture is denoted as such: Eg. 1 - 4x3 - 1: 1 input node, followed by 4 fully-connected layers each having 3 nodes, and 1 output node. For 1D-CNN, the input consists of 3 stock distributions of 100 bins each and all convolutional layers are one-dimensional with a filter width of 3. convf5s2 means a convolutional layer with 5 feature maps and stride 2.}
	\begin{tabular}{cccc}
		\toprule
		& $NLL$ & Model description & No. of parameters \\  \midrule
		\multicolumn{1}{c}{DRN} &  $\mathbf{-474.4(0.1)}$ & No hidden layer  &  7                  \\ 
		\multicolumn{1}{c}{MLP-bins} & -471.5(0.1)    & 300 - 10 - 100 &    4110           \\ 
		\multicolumn{1}{c}{MLP-basis} & -459.8(3.3)  & 45 - 2x5 - 15 &    350            \\ 
		\multicolumn{1}{c}{1D-CNN} & -471.5(0.1)  &  (3x100) - convf5s1 - 2 x convf5s2 - 5 - 100 &  1415       \\ 
		\multicolumn{1}{c}{3BE} &  -466.7(0.7) &  18 basis functions,  450 RKS features &  8100       \\ 
		\bottomrule
	\end{tabular}
	\label{table:stockres}
\end{table*}

\pagebreak
\section*{References}
\bibliography{biblio}

\end{document}